\documentclass[sigconf, nonacm]{acmart}
\usepackage{hyperref}
\usepackage{url}
\usepackage[utf8]{inputenc} 
\usepackage[T1]{fontenc}    
\usepackage{hyperref}       
\usepackage{url}            
\usepackage{booktabs}       
\usepackage{amsfonts}       
\usepackage{nicefrac}       
\usepackage{microtype}      
\usepackage{xcolor}         
\usepackage{multicol}
\usepackage{multirow}
\usepackage{threeparttable}
\usepackage{adjustbox}
\usepackage{array} 
\usepackage{booktabs} 
\usepackage{xspace}
\usepackage{enumitem}
\usepackage{graphicx}
\usepackage{caption}
\usepackage{subcaption}
\usepackage{xcolor}
\usepackage{footnote}
\newcommand{\modelname}{\textsf{VST}\xspace}
\begin{document}
\title{Rec-GPT4V: Multimodal Recommendation with Large Vision-Language Models}

\author[]{Yuqing Liu}
\authornote{Equal contribution}
\affiliation{%
  \institution{University of Illinois at Chicago}
 \city{Chicago}
  \country{United States}
}  
\email{yliu363@uic.edu}

\author{Yu Wang}
\authornotemark[1]
\affiliation{%
  \institution{University of Illinois at Chicago}
 \city{Chicago}
  \country{United States}
}  
\email{ywang617@uic.edu} 

\author{Lichao Sun}
\affiliation{%
  \institution{Lehigh University}
 \city{Bethlehem}
  \country{United States}
}  
\email{lis221@lehigh.edu} 

\author{Philip S. Yu}
\affiliation{%
  \institution{University of Illinois at Chicago}
 \city{Chicago}
  \country{United States}
}
\email{psyu@uic.edu}

\begin{abstract}


The development of large vision-language models (LVLMs) offers the potential to address challenges faced by traditional multimodal recommendations thanks to their proficient understanding of static images and textual dynamics. However, the application of LVLMs in this field is still limited due to the following complexities: First, LVLMs lack user preference knowledge as they are trained from vast general datasets. Second, LVLMs suffer setbacks in addressing multiple image dynamics in scenarios involving discrete, noisy, and redundant image sequences. To overcome these issues, we propose the novel reasoning scheme named Rec-GPT4V: Visual-Summary Thought (\modelname) of leveraging large vision-language models for multimodal recommendation. We utilize user history as in-context user preferences to address the first challenge. Next, we prompt LVLMs to generate item image summaries and utilize image comprehension in natural language space combined with item titles to query the user preferences over candidate items. We conduct comprehensive experiments across four datasets with three LVLMs: GPT4-V, LLaVa-7b, and LLaVa-13b. The numerical results indicate the efficacy of \modelname.
\end{abstract}

\keywords{Large Vision-Language Models, Multimodal Recommendation}

\maketitle

\section{Introduction}

To address the cold-start issues that recommender systems lack sufficient records of new items/users, multimodal recommender systems (MMRSs)~\cite{li2023multimodal, zhou2023attention, wu2022mm, zhou2023bootstrap, he2016vbpr, wei2019mmgcn, zhang2021mining, zhang2023multimodal, liu2023megcf} are proposed by involving the complementary content of items from multiple perspectives, e.g., textual description and visual illustration, thus enriching the recommender system's knowledge. However, the knowledge of MMRSs is primarily learned from scratch using a limited user-item interaction dataset that is often biased and noisy~\cite{chen2019personalized, liu2022elimrec, yang2023modal, zhou2023tale}. 
Additionally, the product image provided by the seller contains critical marketing highlights that attract buyers, e.g., the game’s duration and thematic ambiance, elements that traditional embedding-based MMRSs may struggle to effectively encapsulate.
Moreover, traditional MMRSs encounter challenges in fusing multimodal knowledge, where inefficient integration can further degrade the recommender system's performance~\cite{liu2022disentangled, zhang2022latent, zhou2023comprehensive, liu2023dynamic}.

\begin{figure}
    \centering
    \includegraphics[width=0.6\linewidth]{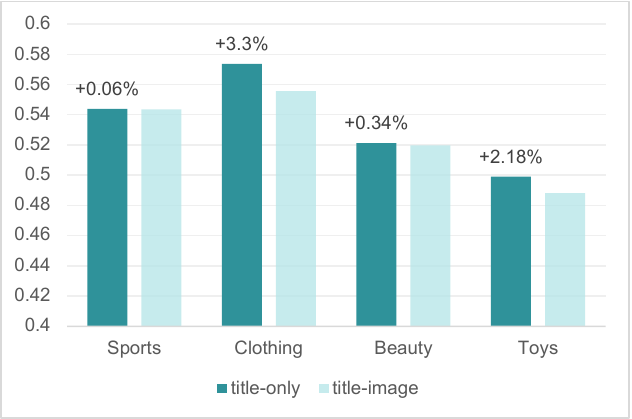}
    \vspace{-1mm}
    \caption{Title-only vs title-image concatenation performance comparison on GPT4-V.}
    \vspace{-5.5mm}
    \label{fig:prelim}
\end{figure}

Meanwhile, the remarkable success of large vision-language models (LVLMs)~\cite{yang2023dawn, wen2023road, zhang2023video, maaz2023video, wang2023gpt4video, li2023blip, zhu2023minigpt, liu2023visual} offers promising solutions to the above issues encountered by traditional MMRSs. LVLMs are proficient in comprehending both textual and visual information about an item, owing to their training on enormous datasets. Their ability to distill and adapt item information across modalities into natural language space exhibits an opportunity for effective knowledge fusion. Despite these strengths, the incorporation of pretrained LVLMs into MMRSs remains an under-explored area. Two possible obstacles may hinder the widespread adoption of LVLMs in MMRSs:

First, \textit{LVLMs are trained from vast general knowledge} and, as such, lack domain-specific knowledge for understanding user preferences revealed through their interactions. This gap leads to LVLMs under-performing compared to traditional MMRSs. To bridge this gap, it is essential to integrate additional knowledge to inform LVLMs in the context necessary for making appropriate recommendations. This approach, however, introduces the second challenge: \textit{LVLMs' inefficiency in processing multiple images}. Although models like GPT4-V have been evaluated in video understanding scenarios to examine their capacity in capturing dynamic content across frames~\cite{yang2023dawn,wen2023road, cao2023towards, tang2023video, wadhawan2024contextual, lu2023foundational, yan2023multimodal}, the scenario with MMRSs involves handling multiple, discrete, and noisy images. This complexity can pose a significant challenge even from a human perspective, making it difficult to extract meaningful knowledge from such diverse interactions. Our preliminary experiments as shown in Figure~\ref{fig:prelim} indicate this issue, showing that a simple concatenation of multiple images with item titles performs worse than methods relying solely on item titles for recommendations even with powerful GPT4-V. Furthermore, current reasoning algorithms, e.g., in-context learning (ICL)~\cite{dong2022survey, rubin2021learning, min2022rethinking, xie2021explanation, li2023finding} and chain-of-thought (CoT)~\cite{wei2022chain, zhang2022automatic, ling2023deductive, wang2022self, trivedi2022interleaving}, are primarily designed for NLP tasks ignoring visual modality. However, the principal challenge in multimodal recommendation is how to effectively leverage image-based knowledge and integrate it into the recommendation process. Thus, effective LVLM-based MMRS requires the design of specific prompting strategies that can utilize their visual comprehension strength without caving to the complexities associated with processing multiple images simultaneously.
\begin{figure*}
\vspace{-25mm}
    \centering
    \includegraphics[width=0.8\textwidth]{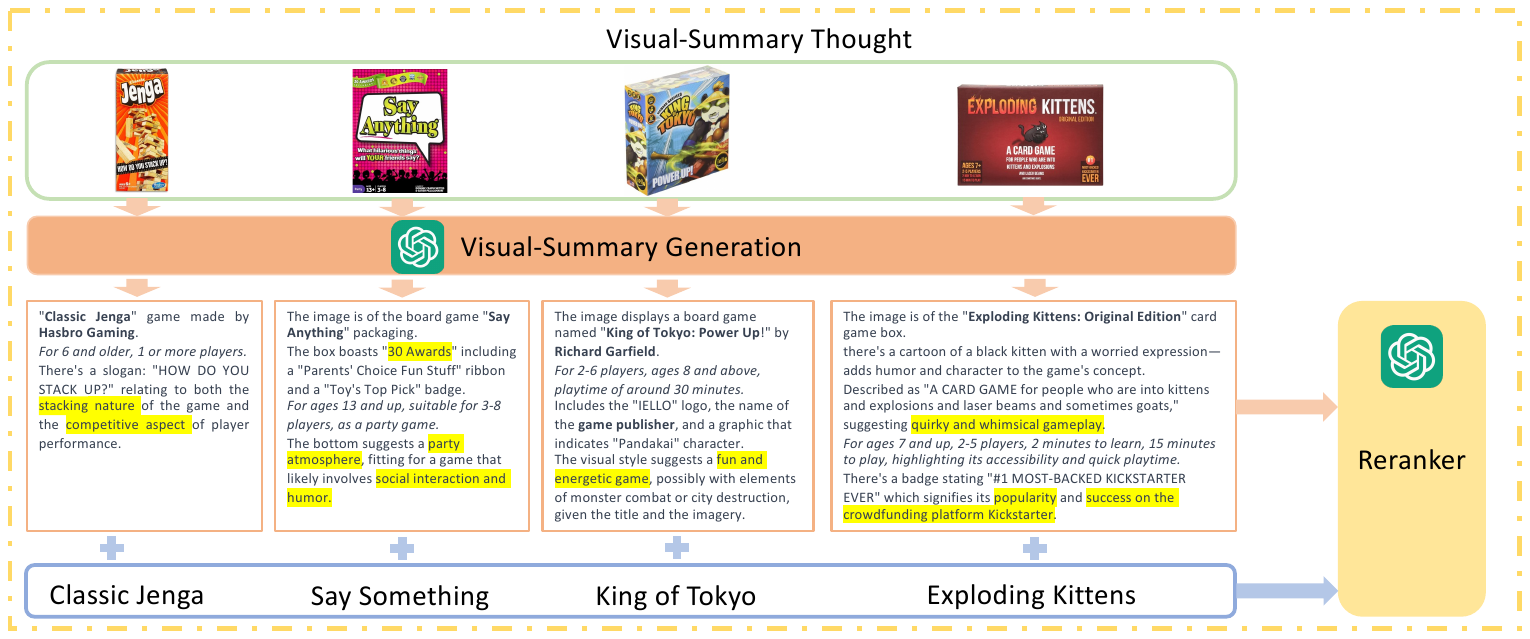}
    \vspace{-30mm}
    \caption{Framework of Visual-Summary Thought of LVLMs for Multimodal Recommendation}
    \label{fig:framework}
\end{figure*}

Accordingly, we propose a novel \textbf{V}isual-\textbf{S}ummary \textbf{T}hought (\modelname) reasoning principle of LVLMs for MMRSs. Our approach includes two primary components: First, we utilize user historical interactions as contextual data for the LVLMs' personalized recommendations. This involves using sequences of both item titles and images as inputs to the LVLMs. Second, to overcome the shortage of handling multiple images, we prompt the LVLMs with one static image to obtain a corresponding summary. Then, we construct user history sequences by substituting the images with their textual comprehensions one by one, serving as an intermediate representation for LVLMs during the reasoning phase. This strategy allows for the recommendation based on a more manageable comprehension of user preferences, transitioning from the complex and noisy image sequences to a simpler task of understanding visual-summary enhanced preference dynamics. To validate the efficacy of our proposed reasoning algorithm, we conducted experiments using GPT4-V, LLaVA-7b, and LLaVA-13b as reasoning backbones. We observe consistent improvements over other existing reasoning strategies, such as concatenation, ICL, and CoT. Our contributions can be summarized as follows:

\begin{itemize}
    \item To the best of our knowledge, this is the first attempt to investigate the reasoning strategies for LVLMs in multimodal recommendation scenarios. 
    \item We introduce a novel Visual-Summary Thought (\modelname) reasoning strategy, specifically designed for the multimodal recommendation context, to harness the proficiency of LVLMs' visual understanding and remedy their deficiency in handling multiple images simultaneously.
    \item We conduct comprehensive experiments to evaluate \modelname, utilizing both API-based LVLMs like GPT4-V, and open-source models such as LLaVA-7b and LLaVA-13b. The consistent improvements observed across these models demonstrate the effectiveness of \modelname for LVLM-based MMRSs.
\end{itemize}
\vspace{-3mm}
\section{Methodology}

\subsection{Problem definition}
In this paper, we follow the problem settings in~\cite{hou2023largellmrank, wang2023drdt} that use the pretrained LVLMs as reranker to recommend the user $u$ via reranking the given $n$ candidate item titles $v = \{v_1, v_2, \dots, v_n\}$. For each user, we have their historical interactions, which is the sequence of title and image pair of items: $u=\{(t_1, i_1), (t_2, i_2), \dots, (t_m,i_m)\}$. 
\subsection{Preliminary}

\paragraph{LVLMs exhibit limitations in handling multiple images} We evaluated the LVLMs' ability to handle multimodal inputs by concatenating the item titles and images of user histories. Surprisingly, leveraging complementary visual information led to poorer results compared to only using item titles as shown in Figure~\ref{fig:prelim}. This underscores a critical insight: adding more information to the LVLMs' prompt context without a thoughtful design can lead to confusion, especially with discrete and noisy images full of redundancy. To address this challenge, we introduce a novel visual-summary thought of prompting strategy (\modelname) as shown in Figure~\ref{fig:framework}.
\vspace{-4mm}
\subsection{Visual-Summary Generation}

Existing LVLMs, e.g., GPT4-V and LLaVA, primarily focus on static image understanding scenarios, where LVLMs generate textual descriptions of a given image. However, this paradigm is inefficient for handling multiple images~\cite{wen2023road}. Existing strategies include concatenating images for LVLM reasoning~\cite{wen2023road}, or adapting LVLMs to video comprehension scenarios via finetuning on video datasets~\cite{maaz2023video, zhang2023video, wang2023gpt4video, tang2023video}. Yet, neither approach is suitable for the unique demands of MMRSs, where the image sequence of a user history is discrete and noisy, lacking the continuous nature of video frames and making sequential correlations difficult to discern. To deal with these issues, we propose leveraging LVLMs' strengths in temporal understanding within natural language modality and their capacity for static image interpretation. Instead of processing a sequence of images, we focus on distilling critical marketing highlights from individual image. The prompt can be formalized as:
$s_{i} = sumary(i)=$\textit{"What's in this image?"} In this way, we can not only obtain marketing highlights of items via distilling image comprehension from LVLMs but also simplify the temporal user preference understanding from the visual modality to the textual modality, where the LVLMs demonstrate proficiency. 
\vspace{-3mm}
\subsection{Visual-Summary Thought for MMRSs}

After summarizing each item image, we concat the history item titles with their visual summary to construct the prompt for querying user preferences among candidates. The prompt is structured in two parts: the first outlines the user's purchase history in chronological order, demonstrated by each item's title and visual summary. The second segment directs the LVLMs to rerank the candidates represented by their titles. An illustrative prompt might be: 

\textit{"[Here is a chronological list of my purchase history for some products including the title and the description of each product. $\{(t_1, s_{i_1})$, \dots, $(t_m, s_{i_m})\}$][There are $|n|$ candidate products I am considering to buy: $\{v_1, \dots, v_n\}$. Please rank these $|n|$ candidate products based on the likelihood that I would like to purchase next most according to the given purchase history. You cannot generate products that are not in the given candidate list.]".}
\begin{table*}[!ht]
    \caption{Performance comparison of different prompt strategies. Target items are guaranteed to be included in the candidate sets. We highlight the \textbf{best} and the \underline{second-best} results.}
    \vspace{-3mm}
    \label{tab:exp-overall_comp}
    \small
    \centering
    \begin{adjustbox}{width=0.8\textwidth}
    \begin{threeparttable}
    {
    \begin{tabular}{l|l|cccc|cccc|cccc}
         \toprule
         \multirow{2}{*}{\textbf{Dataset}} 
         &  \multirow{2}{*}{\textbf{Metric}} & \multicolumn{4}{c|}{\textbf{GPT4V}} & \multicolumn{4}{c|}{\textbf{LLaVA-7b}} & \multicolumn{4}{c}{\textbf{LLaVA-13b}}  \\
         &  &  MM & MM-ICL & MM-CoT  & \modelname & MM & MM-ICL & MM-CoT  & \modelname &MM & MM-ICL & MM-CoT  & \modelname \\
         \midrule
        \multirow{6}{*}{\textbf{Sports}}
         & R@5  & 0.6900 & \underline{0.695}0 & 0.5750 & \textbf{0.7250} &0.1300 & \underline{0.1900} & 0.1800 & \textbf{0.3283} & 0.2250 & \underline{0.3300} & 0.2300 & \textbf{0.3750}  \\
         & R@10 & \underline{0.8600} & \underline{0.8600} & 0.8150 & \textbf{0.9000} &  0.2950 & \underline{0.3400} & 0.3250 & \textbf{0.5067} & 0.3200 & \underline{0.4850} & 0.3250 & \textbf{0.6250}  \\
         & R@20 & \underline{0.8700} & 0.8650 & 0.8300 & \textbf{0.9050} & 0.3100 & 0.3500 & \underline{0.3550} & \textbf{0.5117} & 0.3400 & \underline{0.5000} & 0.3450 & \textbf{0.6350} \\
         & N@5  & 0.4880 & \underline{0.5126} & 0.4186 & \textbf{0.5263} & 0.0703 & \underline{0.1138} & 0.1043 & \textbf{0.1769} & 0.1395 & \underline{0.2087} & 0.1393 & \textbf{0.2244} \\
         & N@10 & 0.5435 & \underline{0.5666} & 0.4961 & \textbf{0.5834} & 0.1243 & \underline{0.1619} & 0.1506 & \textbf{0.2345} & 0.1706 & \underline{0.2598} & 0.1701 & \textbf{0.3063} \\
         & N@20 & 0.5461 & \underline{0.5678} & 0.4999 & \textbf{0.5846} & 0.1281 & \underline{0.1646} & 0.1580 & \textbf{0.2357} & 0.1755 & \underline{0.2637} & 0.1752 & \textbf{0.3086} \\
         \midrule
        \multirow{6}{*}{\textbf{Clothing}}
         & R@5  & 0.6550 & \textbf{0.7100} & 0.6300 & \underline{0.6950} & 0.1400 & 0.1650 & \underline{0.1700} & \textbf{0.2800} & \underline{0.3650} & 0.3200 & 0.2550 & \textbf{0.3950} \\
         & R@10 & 0.8950 & \underline{0.9050} & 0.8150 & \textbf{0.9300} & 0.2750 & \underline{0.3100} & 0.2600 & \textbf{0.3250} & \textbf{0.6700} & 0.5450 & 0.4200 & \underline{0.6200}  \\
         & R@20 & 0.9000 & \underline{0.9050} & 0.8200 & \textbf{0.9350} & 0.2900 & \underline{0.3150} & 0.2600 & \textbf{0.3250} & \textbf{0.6950} & 0.5450 & 0.4200 & \underline{0.6250} \\
         & N@5  & 0.4781 & \textbf{0.5580} & 0.4631 & \underline{0.5322} & 0.0851 & \underline{0.1156} & 0.1086 & \textbf{0.1875} & \underline{0.2248} & 0.2062 & 0.1554 & \textbf{0.2594 } \\
         & N@10 & 0.5555 & \textbf{0.6205} & 0.5238 & \underline{0.6085} & 0.1287 & \underline{0.1633} & 0.1386 & \textbf{0.2025} & \underline{0.3234} & 0.2787 & 0.2058 & \textbf{0.3329} \\
         & N@20 & 0.5569 & \textbf{0.6205} & 0.5252 & \underline{0.6098} & 0.1326 & \underline{0.1646} & 0.1386 & \textbf{0.2025} & \underline{0.3301} & 0.2787 & 0.2085 & \textbf{0.3343} \\
         \midrule
                 \multirow{6}{*}{\textbf{Beauty}}
         & R@5  & \textbf{0.6300} & \textbf{0.6300} & 0.5500 & 
         \underline{0.6200} & \underline{0.2450} & 0.1800 & 0.1450 & \textbf{0.2750} & 0.2650 & \underline{0.2900} & 0.2300 & \textbf{0.3200} \\
         & R@10 & 0.8450 & \underline{0.8700} & 0.6400 & \textbf{0.9000}& \textbf{0.4050} & 0.3150 & 0.1700 & \underline{0.4000} & 0.3750 & \underline{0.4200} & 0.3200 & \textbf{0.5500} \\
         & R@20 & 0.8500 & \underline{0.8750} & 0.6500 & \textbf{0.9000} & \textbf{0.4200} & 0.3200 & 0.1750 & \underline{0.4000} & 0.3850 & \underline{0.4200} & 0.3250 & \textbf{0.5600} \\
         & N@5  & \underline{0.4503} & 0.4395 & 0.3964 & \textbf{0.4536} & \underline{0.1484} & 0.1202 & 0.1006 & \textbf{0.1769} & 0.1641 & \underline{0.1928} & 0.1398 & \textbf{0.2183}\\
         & N@10 & \underline{0.5197} & 0.5183 & 0.4264 & \textbf{0.5439} & \underline{0.1996} & 0.1641 & 0.1087 & \textbf{0.2179} & 0.2008 & \underline{0.2361} & 0.1692 & \textbf{0.2942} \\
         & N@20 & \underline{0.5211} & 0.5195 & 0.4290 & \textbf{0.5439} & \underline{0.2035} & 0.1655 & 0.1101 & \textbf{0.2179} & 0.2033 & \underline{0.2361} & 0.1706 & \textbf{0.2970}  \\
         \midrule
                 \multirow{6}{*}{\textbf{Toys}}
         & R@5  & 0.5500 & \textbf{0.6450} & 0.4950 & \underline{0.6300} & \underline{0.1450} & 0.1150 & 0.1300 & \textbf{0.3000} & 0.1875 & \underline{0.3400} & 0.2600 & \textbf{0.3617} \\
         & R@10 & 0.7650 & \underline{0.7800} & 0.6950 & \textbf{0.8000} & \underline{0.2750} & 0.1450 & 0.1700 & \textbf{0.3800} & 0.2550 & \underline{0.4250} & 0.3800 & \textbf{0.5150} \\
         & R@20 & 0.7750 & \underline{0.7800} & 0.7050 & \textbf{0.8000} & \underline{0.2850} & 0.1550 & 0.1850 & \textbf{0.3950} & 0.2663 & \underline{0.4350} & 0.3800 & \textbf{0.5200} \\
         & N@5  & 0.4184 & \textbf{0.4789} & 0.3967 & \underline{0.4399} & \underline{0.0857} & 0.0842 & 0.0835 & \textbf{0.2035} & 0.1389 & \underline{0.2373} & 0.1832 & \textbf{0.2412} \\
         & N@10 & 0.4883 & \textbf{0.5227} & 0.4349 & \underline{0.4958} & \underline{0.1281} & 0.0941 & 0.0977 & \textbf{0.2299} & 0.1614 & \underline{0.2648} & 0.2228 & \textbf{0.2919} \\
         & N@20 & 0.4911 & \textbf{0.5227} & 0.4376 & \underline{0.4958} & \underline{0.1305} & 0.0966 & 0.1015 & \textbf{0.2336} & 0.1642 & \underline{0.2672} & 0.2228 & \textbf{0.2932} \\
         \bottomrule
    \end{tabular}
    }
    \end{threeparttable}
    \end{adjustbox}
\end{table*}

\section{Experiments}
In this section, we provide the performance comparison between the proposed \modelname and three representative reasoning strategies on four public datasets, using GPT4-V, LLaVA-7b, and LLaVA-13b as pre-trained LVLMS.

\subsection{Experimental Settings}
\paragraph{Dataset}
In this paper, we adopt the same dataset as in~\cite{geng2023vip5} that uses the Amazon Review datasets for evaluation. Due to the limitation of the inference rate, following the common practice~\cite{hou2023largellmrank}, we only sample 200 users for evaluation. We report the statistics of such datasets in Table~\ref{tab:data_stats}.
\begin{table}[]

    \centering
    \caption{Statistics of the datasets after sampling.}
    \vspace{-3mm}
    \begin{adjustbox}{width=0.9\columnwidth}
    \begin{tabular}{lcccc}
    \toprule
         Datasets & \#Users & \#Items & \#Interactions & Sparsity \\
    \midrule
        Sports & 200 & 1750 & 2333 & 99.33\% \\
        Clothing & 200 & 1291 & 1362 & 99.47\% \\
        Beauty & 200 & 2024 & 2797 & 99.31\%\\
        Toys & 200 & 1684 & 1967 & 99.42\% \\
    \bottomrule
    \end{tabular}   
        \end{adjustbox}
    \label{tab:data_stats}
    \vspace{-3mm}
\end{table}
\paragraph{Metrics} We adopt Recall@K (R@K) and NDCG@K (N@K) to evaluate the ranking performance of the LVLMs over candidate items, which consist of the title of the ground-truth (target) item and the 9 random sampled items.
\paragraph{Implementation Details} For open-source LVLMs, we use Fastchat to launch models and conduct the model inference on a single GeForce RTX 4090. 

\paragraph{Baseline Models}
As there is no previous work that only utilizes the inference capacity of LVLMs for multimodal recommendation, we adopt the commonly chosen prompting strategies used in NLP tasks: in-context-learning and chain-of-thought for comparison. For ICL, we match each prefix of the user's historical interaction sequence with its corresponding successor as demonstration examples. For example: \textit{"[Here is a chronological list of my purchase history: $\{(t_1, s_1)$, \dots, $(t_{m-1}, s_{m-1})\}$] [Then if I ask you to recommend a new product, you should recommend $t_m$. Now I've just purchased $t_m$, I want to buy a new product...]"}. The remaining part is the same as the second part of \modelname. For CoT, we adopt zero-shot CoT by adding \textit{"Please think step by step."} to the second part of the prompt. Furthermore, we also compare our model with the plain prompt, which is the concatenation of the historical item titles and images. 
\begin{figure}
    \centering
    \includegraphics[width=0.9\linewidth]{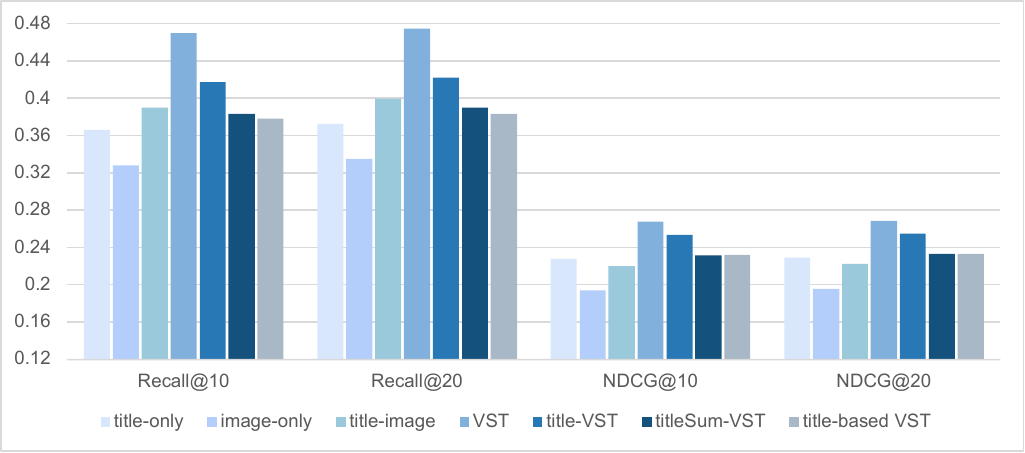}
    \vspace{-2mm}
    \caption{Ablation study. Performance of LLaVA-13b with different prompts on Toys dataset.}
    \label{fig:ablation}
    \vspace{-5mm}
\end{figure}
\begin{figure*}
    \centering
    \includegraphics[scale=0.4]{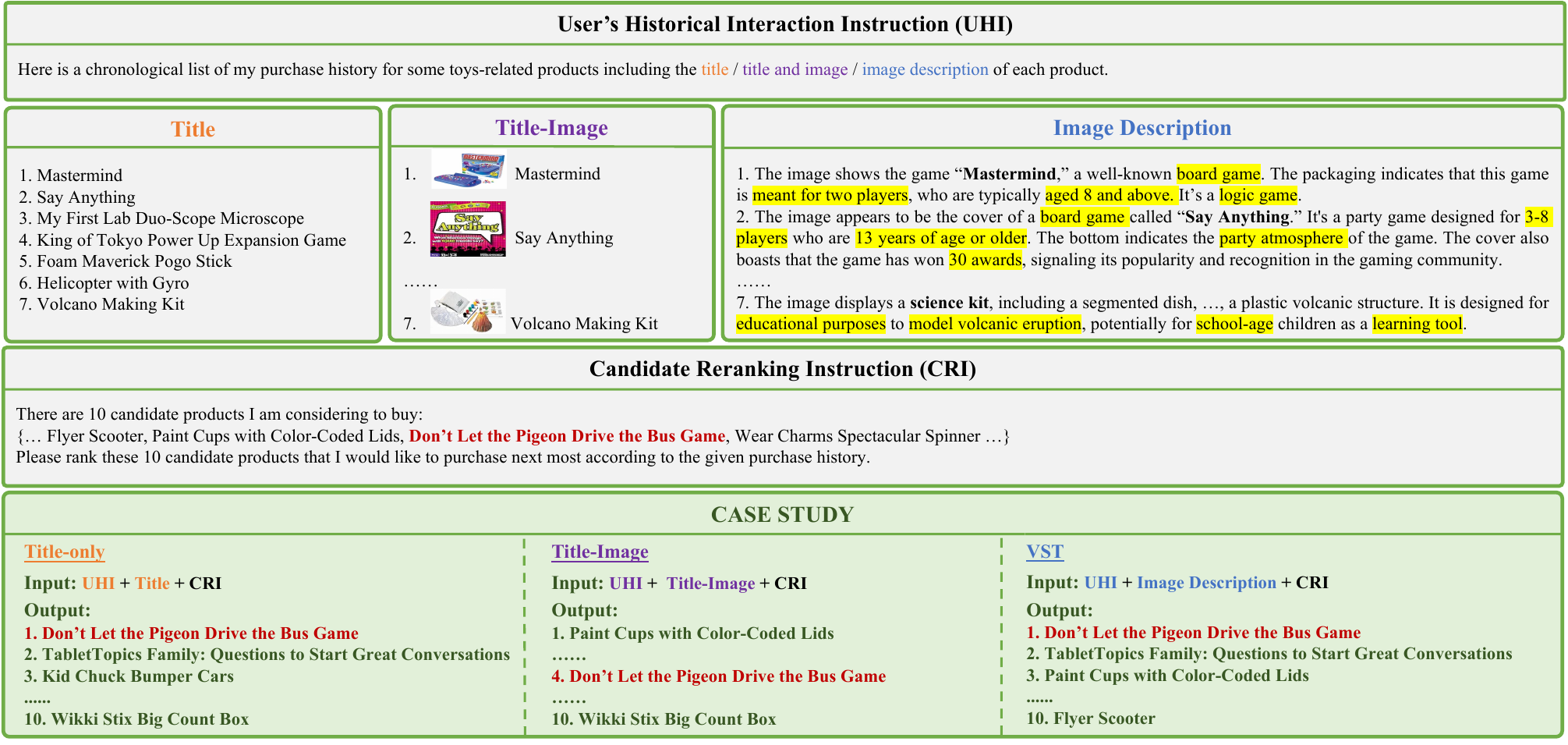}
    \vspace{-3mm}
    \caption{Case study. Text in red indicates the target item. Text in orange, purple, or blue indicates the pattern to describe the item for the corresponding prompt. Text in yellow highlights some key features obtained through visual-summary generation. }
    \label{fig:case study}
\end{figure*}
\subsection{Overall Performance}
To demonstrate the effectiveness of our proposed \modelname strategy, we employ GPT4-V, LLaVA-7b, and LLaVA-13b as pre-trained LVLMS and conduct experiments with four different prompt strategies across four datasets. The complete experimental results are shown in Table~\ref{tab:exp-overall_comp}. 
From the table, we can observe that our proposed \modelname reasoning strategy achieves the best or comparable performances across all datasets, demonstrating the effectiveness of our approach. Notably, our approach has a better performance on Sports dataset than others. This might be due to the titles of this category of products containing much more noise, making the alignment between textual and visual information more challenging for the employed LVLMs. In contrast, through visual-summary generation, \modelname can capture more relevant information from the image and can reduce the impact of the noise in the title to some extent.

\subsection{Ablation Study}
To analyze the effectiveness of the \modelname reasoning principle, we conduct an ablation study on six variants of the proposed strategy. The results on Toys dataset using LLaVA-13b are shown in Figure~\ref{fig:ablation}. The reported results are the average of a minimum of three repeated runs, aimed at minimizing the impact of randomness.
\textit{titleSum-\modelname} refers to the prompt that also lets LVLMs distill information from the title of an item: $s_t=summary(t)=$\textit{"What information can you get from the title?"}, then appended by the summary distilled from the corresponding image. \textit{title-based \modelname} refers to instructing LVLMs to distill information from an image by taking item title into consideration, where $s_i=summary(i)=$\textit{"This is an image related to $t$. Please provide a detailed description of the given image."}

From the results, we have the following observations: (1) \modelname can capture more meaningful information from both textual and visual modalities. The results show that \modelname has the capability to significantly enhance the ranking performance compared to non-\modelname-based strategies. The improvement stems from \modelname's proficiency in multimodal understanding and serves better in sequential scenarios, where information from different sources needs to be integrated effectively. (2) Information from the title can boost performance, but it depends on the quality of the title and the alignment between the title and the image. Compared to the results among \modelname, title-\modelname, titleSum-\modelname, and title-based \modelname, we can observe that adding the title information doesn't yield improvement. 
This lack of improvement is likely due to the visibility of toy titles in images or the easy identification of entities mentioned in titles from the images themselves. Therefore, combining title information with \modelname does not provide substantial additional benefits. Whether to include titles during reasoning remains a hyperparameter decision dependent on the quality of titles in each dataset.

\subsection{Case Study}
In this section, we compare the ranking lists generated by LLaVA-13b using \modelname with title-only and title-image concatenation prompts. The results are shown in Figure~\ref{fig:case study}. Here are our observations from comparing the outputs: Both title-only and \modelname strategies successfully rank the target item as the first position, while the naive concatenation of title and image places it fourth. This discrepancy suggests that raw images may contain an excess of information, which could be perceived as redundant and introduce additional noise into our ranking task. On the other hand, the \modelname strategy offers a more refined approach. By utilizing \modelname, we not only incorporate information from the title but also extract richer and more relevant details from the image itself. Such details also align closely with the marketing selling points of the product. Consequently, the \modelname strategy emerges as a more effective prompt for multimodal recommendation, as it combines textual and visual cues to provide a comprehensive understanding of the item, thereby enhancing the performance of the ranking results.


\section{Conclusion}
In this work, we investigate the performance of different reasoning strategies for LVLMs in multimodal recommendation scenarios and identify a notable limitation in LVLMs' capability to effectively handle multiple images. To bridge this gap, we propose a Visual-Summary Thought (\modelname) strategy to distill information from images. By leveraging LVLMs' visual understanding, \modelname aims to harness their strengths while rectifying deficiencies in handling multiple images. Extensive experiments conducted on four real-world datasets using both API-based LVLMs such as GPT4-V and open-source models like LLaVA-7b and LLaVA-13b, consistently demonstrate the effectiveness of our proposed strategy.

\bibliographystyle{ACM-Reference-Format}
\bibliography{reference}
\end{document}